\documentclass[sn-mathphys,Numbered]{sn-jnl}

\usepackage{graphicx}%
\usepackage{multirow}%
\usepackage{amsmath,amssymb,amsfonts}%
\usepackage{amsthm}%
\usepackage{mathrsfs}%
\usepackage[title]{appendix}%
\usepackage{xcolor}%
\usepackage{textcomp}%
\usepackage{manyfoot}%
\usepackage{booktabs}%
\usepackage{algorithm}%
\usepackage{algorithmicx}%
\usepackage{algpseudocode}%
\usepackage{listings}%
\usepackage{subfig}

\theoremstyle{thmstyleone}%
\theoremstyle{thmstyletwo}%

\theoremstyle{thmstylethree}%

\begin{document}

\title[Article Title]{Distributed client selection with multi-objective in federated learning assisted Internet of Vehicles}


\author*[1]{\sur{Narisu Cha}}\email{nrs018@gmail.com}

\author[2]{\sur{Long Chang}}\email{changlong@imufe.edu.cn}
\equalcont{These authors contributed equally to this work.}

\affil*[1]{\orgdiv{The School of Computer and Information Management}, \orgname{Inner Mongolia University of Finance and Economics}, \orgaddress{\street{North second ring No. 185}, \city{Huhhort}, \postcode{010070}, \state{Inner Mongolia }, \country{China}}}
\affil[2]{\orgdiv{The School of Statistics and Mathematics}, \orgname{Inner Mongolia University of Finance and Economics}, \orgaddress{\street{North second ring No. 185}, \city{Huhhort}, \postcode{010070}, \state{Inner Mongolia }, \country{China}}}


\abstract{Federated learning is an emerging distributed machine learning framework in the Internet of Vehicles (IoV). In IoV, millions of vehicles are willing to train the model to share their knowledge. Maintaining an active state means the participants must update their state to the FL server in a fixed interval and participate to next round. However, the cost by maintaining an active state is very large when there are a huge number of participating vehicles. In this paper, we proposed a distributed client selection scheme to reduce the cost of maintaining the active state for all participants. The clients with the highest evaluation are elected among the neighbours. In the evaluator, four variables are considered including sample quantity, throughput available, computational capability and the quality of the local dataset. We adopted fuzzy logic as the evaluator since the closed-form solution over four variables does not exist. Extensive simulation results show our proposal approximates the centralized client selection in terms of accuracy and can significantly reduce the communication overhead.}




\maketitle

\section{Introduction}
Over the past years, federated learning assisted Internet of vehicles \cite{pmlr_v54_mcmahan17a,FL2021advances, FLinIoV2021IEEEnetwork, FLinIoV2} have been received widespread attention from academia and industry because federated learning as a distributed machine learning framework, can achieve the purpose of exchanging knowledge cross the user through sharing training models and avoid the leakage of the privacy, without uploading raw data related to user privacy to the server. In the framework, millions of client devices owning local datasets collected by terminal equipment are willing to participate in training. For example, in the GBoard (a keyboard for tablets and smartphones developed by Google Inc.), over 1.5 million clients are chosen for the learning language model in the whole process of federated training \cite{gboard}. Also, in FL-assisted vehicular networks, there are over 3 million vehicles in the active state in the Kanto area, Japan. Because of the wireless bandwidth limitation, a small fraction of the users are only chosen as the client for learning in each round. The server is responsible for client selection in the classical federated learning framework. Hence, all participants willing to train the model must update their state as alive in the fixed interval. In general, as mentioned in \cite{gboard}, the user device must meet some conditions, including dataset freshness and size, charging, idle wireless connection, operating system version and hardware requirements. In general, in the FL-assisted vehicular networks, the vehicle states can include the location, network environment and running state, such as the velocity and the acceleration. These states need to update the server over wireless connection promptly. Compared to the uploading model, the overhead of updating the state from all participants to the server is not negligible since millions of participants need to update the state at any time.

Most classical federated learning frameworks adopt a centralized control scheme when the server selects the clients for the learning. The server is responsible for collecting and updating the active state of all participants. And then, the server chooses a set of participants with active states as training clients at the beginning of a new communication round. Undoubtedly, the overhead that occurred by updating the state is huge while the number of participants is tremendous. These overheads occupy precious wireless resources and increase communication delays, and even most are useless. As the conclusion in work \cite{fedcs2019}, selecting a small fraction of devices in each round can also achieve good performance. Unfortunately, most researchers did not consider the overhead from updating the state of all participants. They only proposed a framework adapting decentralized client selection by updating the active state of all participants.

The server has to transfer the authority of selecting the clients to the clients because the server in the distributed framework will no longer accept the message about the participants for the evaluation. Meanwhile, the assessment of all participants is moved from the server to the local. Each participant assesses himself according to the input parameters, such as sample quantity, network throughput, computational capability and the diversity of the local dataset. 

The fundamental differences between centralized client selection and distributed client schemes are described as follows. The first difference is whether the information of all participants for evaluation is collected on the server side. The second difference is whether the server has the power to select clients. Furthermore, assessing the participant is processed locally instead of on the server. Most of the traditional federated learning frameworks are developed on the centralized scheme. Hence, the overhead of updating information is very large and can not be eliminated. On the contrary, the distributed framework with client selection can eliminate the overhead smoothly.

In the federated learning-assisted vehicular networks, selecting some vehicles with good performances in each round can effectively speed up the convergence of the model. For example, a client with a larger training dataset, with stronger computational capability, higher network throughput, and the loss function of the local dataset can ensure that the client model is successfully uploaded to the central server. Thus, global model accuracy is improved while more local models participate in aggregation. However, a close-form solution considering the influencing factors does not exist and brings a huge challenge to evaluating the participating vehicle. To this end, we adopt a lightweight evaluating approach, fuzzy logic based client evaluator, that can utilize the fuzzy relationship between influencing factors to construct an assessment approach for the participants. 

This paper aims to eliminate the communication overhead between a huge amount of participants and the server to maintain the active state of the participants on the server side. The main contributions of the paper are listed as follows.

\begin{itemize}
    \item A radically different framework, named distributed client selection framework, is proposed in which the server is not in charge of the client selection process and without gathering the information of all participants to eliminate the communication overhead for keeping a huge amount of participants active state. 
    \item A client evaluator with multi-objective, named fuzzy evaluator, is proposed to assess a client for the contribution of the model convergence. In the evaluator, four objectives are considered, specifically, sample quantity on the local, available throughput and computational capability as well as loss function of the local dataset. Moreover, the evaluator has to move from the central server to the participant because the server would not collect the client's information.
   
    \item To verify our framework, a simulator with realistic vehicular network and distributed machine learning is constructed. Meanwhile, a non-i.i.d dataset with different levels is synthesized to test the superiority of our framework when the dataset has heterogeneity. 
\end{itemize}

The remainder of the paper is organized as follows. In Section \ref{section2}, some significant existed works are summarized. In Section \ref{section3}, federated learning assisted Internet of vehicular is presented. And then, the distributed client selection framework and the client evaluator with multi-objective are elaborated in detail in Section \ref{section4} and Section \ref{section5}, respectively. And, in Section \ref{section6}, the simulator is set up and the experimental results are discussed. Finally, the conclusion is presented in Section \ref{section7}.

\section{Related Work}
\label{section2}

Over the past years, researchers paid attention to a bottleneck in the communication cost between the server and the client for exchanging models in the FL \cite{pmlr-v119-hamer20a, KonecnyMYRSB16, wang2019icdcs, niknam2020federated}. To mitigate the communication cost, researchers proposed some solutions \cite{2023suervey}, including gradient and model compression \cite{pmlr-v119-hamer20a, KonecnyMYRSB16}, biased client selection \cite{wang2019icdcs, pmlr-v151-jee-cho22a, fedcs2019}, reinforcement learning based client selection \cite{RLbasedclientselectionINFOCOM2020, RLclientselection2021}, learning on edge \cite{niknam2020federated} and so on. So far, all federated learning framework is based on centralized client selection in which the FL server is in charge of the client selection process and selecting information about all participants must be gathered by the FL server to maintain the active state of the participants. In fact, millions of participants are willing to the training in the federated learning. Hence, the communication overhead caused by maintaining the active state for all participants is \textbf{FAR LARGER} than the communication overhead caused by exchanging the model between the FL server and the client. Unfortunately, most researchers did not pay attention to the overhead by maintaining the active state. Although, hierarchical federated learning \cite{hierarchical2023} can mitigate the communication overhead from maintaining the active state of all participants, the centralized client selection framework does not eliminate the overhead. In regret that the authors in \cite{hierarchical2023} did not consider the overhead that the participant maintains the active state. 

The decentralized federated learning (DFL) framework is widely discussed, in which aggregating server does not need, and the training model of each client is sent to all other clients. The authors in \cite{decentralizedFL2019arXiv} considered a DFL based on peer-to-peer communication to serve medical applications. The authors in \cite{decentralizedIEEEnetwork2020} considered a DFL framework using the committee consensus blockchain and all models, including the global model from the server and the local model from the client are stored in the blockchain. In the same way, the authors in \cite{blockchainFLCS2020} also considered an asynchronous DFL framework based on blockchain to enhance the stability and reliability during model transmission in the IoV. The work \cite{blockchain2020decentralized} proposed decentralized federated learning based on blockchain for the vehicular network and analyzed the advantages from the perspective of the theory. In addition, the authors in \cite{decentralized2022} developed an approach, which updates the model according to the connection state, even partially received the model, to suit unreliable networks. The DFL framework mentioned above is very suitable for the dynamics network environment, such as vehicular networks with mobility. However, the size of exchanging model among all clients increased exponentially with the number of participants. For example, exchanging model size is very huge in the IoV owning millions of vehicles. In addition, these DFL frameworks only consider uploading models between all clients, and the communication cost occurred by maintaining the active state of all participants needs to be addressed.
 
 Different from the centralized client selection framework, in the distributed client selection, the server would not gather all participant information. And the client evaluation /selection is removed out from the server. For the client selection, it can be classified as biased client selection and unbiased client selection, respectively. Unbiased client selection does not consider any factors, for example, random selection \cite{fedcs2019}, in which all participants have the same chance to be selected as the client. In contrast, biased client selection selects the client according to some sorting. The authors in \cite{fu2023clientselection} systematically summarized the opportunities and challenges in the client selection process and highlighted the importance of system and data heterogeneity to client selection. The authors in \cite{FL2020survey} investigated most of the existing works involving system architecture, application, privacy concerns as well as resource management. The authors in \cite{fairnessCS2023} summarized the taxonomy and challenges of the client selection in terms of fairness to create an incentive for the sustainability of the FL ecosystem. Similarly, the authors in \cite{fairfed2023ezzeldin} developed a novel global model aggregation algorithm, which considered group fairness instead of the weight related to the sample quantity in the local. The authors in \cite{pmlr-v151-jee-cho22a} choose the client with a larger loss function to speed up the convergence. The authors in \cite{xu2021transwirelcom} jointly consider wireless resource allocation and client selection from a long-term perspective. The authors in \cite{multicriteriaclientselection2021} consider multiple criteria, such as computational capability, memory and energy, in the client selection to maximize the successful ratio of the uploading model. The authors in \cite{blockchainFLCS2020} proposed an asynchronous federated learning, in which deep reinforcement learning (DRL) resided on the server selects the nodes with higher communication and computation resources for the training. The local model is uploaded to the blockchain instead of the server. In the same way, the global model is also distributed to the blockchain. Finally, in the aggregation, the server retrieves from the blockchain to aggregate the global model after the local training. The authors in \cite{efficiency2021} considered the client selection from efficiency and fairness. In complex networks, such as federated learning assisted Internet of vehicular, the approaches mentioned above still faced challenges because of the heterogeneity of the client. To this end, the authors in \cite{fuzzyCentralized2022} proposed a client selection considering multi-objective evaluation. Unfortunately, the work \cite{fuzzyCentralized2022} is based on the centralized client selection framework and did not eliminate the communication overhead. 

The heterogeneity among clients is a key challenge in federated learning, including statistical and system heterogeneity. The heterogeneity not only dropped the accuracy but also slowed the convergence speed. To address the statistical heterogeneity, FedProx \cite{fedproxMLSYS2020} added a proximal term to the local objective function to reduce the gradient drift. However, the authors in \cite{fedproxMLSYS2020} did not consider the system heterogeneity. For example, a connection is broken by vehicle mobility in IoV. The authors in \cite{wang2021lossfunction} jointly considered node selection and wireless resource allocation in the heterogeneous FL system to maximize loss function decay and accelerate the convergence. To address the non-i.i.d of the dataset, the authors in \cite{featuredetection2023IoT} adopted a support vector machine (SVM) to detect the feature of samples and remove the useless samples. SCAFFOLD \cite{scaffold2020pmlr} corrected the direction of the update by the difference between the global model and the local model. Prior works provide important references in terms of the theory and method. Unfortunately, these works lack the combination with the real world.

\section{Federated learning assisted IoV}
\label{section3}

Considering the critical challenges of intelligent transportation systems (ITS) are that system heterogeneity, model performance and user privacy \cite{FLinIoV1}. System heterogeneity refers to the resources, such as computational capability, available network resources, as well as training datasets owned by each vehicle running on the road, which differ from other vehicles. In general, these resources vary with the vehicle state as well. For example, a vehicle parking on the lot has more resources compared to a vehicle fast driving on the highway in terms of computational capability and training dataset. In the same way, available network resource varies with vehicle movement, while the vehicle move in and out continuously from the signal coverage of the roadside unit (RSU) located aside the road. The model performance worsens drastically when the network environment changes with the vehicle's mobility. The model having stable and high safety is important for safeguarding passengers and pedestrians. The privacy concern for the vehicle, such as trajectory and traffic context, not only affects the driving experience, it can even endanger the life of pedestrians. 

\subsection{Federated learning}

Federated learning is a distributed machine learning paradigm in which user data are kept local during the learning to protect the user's privacy concern.
In the classical federated learning \cite{pmlr_v54_mcmahan17a}, the whole process in each round is divided into four steps: broadcast global model, local updates over local data, uploading local model and aggregation global model for the next round. 

Each client updates their local model over local data according to the following Equation \ref{equ4_3}:

\begin{equation}
\label{equ4_3}
    \textbf{w}_{i}^{k+1} = \textbf{w}_{i}^{k} - \eta * \frac{\partial L_{i}(\textbf{w}_{i}^{k})}{\partial \textbf{w}_{i}^{k}},
\end{equation}
where $\eta$ is referred to as the learning rate. $\textbf{w}_{i}^{k}$ and $L_{i}(\textbf{w}_{i}^{k})$ are referred to local model and the loss function of the client \textit{i}, respectively. The local model is uploaded to the server for aggregation after the training.

The global model for the next round is generated on the server by the following Equation \ref{equ4_4}:

\begin{equation}
\label{equ4_4}
    \textbf{w}_{g}^{k+1}=\sum_{i=1}^{N}\frac{\lvert D_{i}^{k} \rvert}{\lvert D^{k} \rvert}\textbf{w}_{i}^{k}.
\end{equation}

And global loss function is defined by the following Equation \ref{GlobalLoss}:

\begin{equation}
\label{GlobalLoss}
    L_{g}(\textbf{w}_{g}^{k+1})=\sum_{i=1}^{N}\frac{\lvert D_{i}^{k} \rvert}{\lvert D^{k} \rvert}L_{i}(\textbf{w}_{i}^{k}),
\end{equation}
where \textit{N} refers to the number of selected clients in the round \textit{k}. And $\lvert D^{k} \rvert = \sum_{i} \lvert D_{i}^{k} \rvert$. $\textbf{w}_{g}^{k+1}$ is referred to the global model in the round \textit{k+1}. The categorical cross-entropy loss is adopted to output the probability of the multi-classification in the local training.

\begin{equation}
\label{equ4_2}
    \min_{\textbf{w}} L_{g}(\textbf{w})=\min_{\textbf{w}}\sum_{i=1}^{N}\frac{\lvert D_{i}^{k} \rvert}{\lvert D^{k} \rvert}L_{i}(\textbf{w}_{i}^{k}).
\end{equation}

\subsection{Federated learning assisted IoV}
Federated learning is realized as an emerging effective manoeuvre for protecting user privacy and can be applied to the IoV regarding data sharing, collaborative intelligence and distributed machine learning \cite{datasharingFL, blockchainFLCS2020}. In the future, the vehicle with intelligence will be mainstream in the ITS, in which deployed various AI apps to assist driving. Additionally, the vehicle has some resources such as computational capability, communication ability as well as storage for processing the data collected by the sensors. These vehicles not only exchange basic information with each other, but also share the knowledge learned from the AI model among the vehicles.   

\section{Distributed client selection framework}
\label{section4}

In this section, we briefly describe the difference between centralized federated learning (CFL) and distributed federated learning (DFL), as well as three different client selection schemes. And then, two kinds of communication overhead in federated learning, specifically, the overhead by exchanging models between the FL server and the clients, and the overhead by maintaining the active state of all participants, would be presented. Next, we compare two kinds of overhead using GBoard. Finally, the distributed client selection is described in detail.

\subsection{Different client selection schemes}

In the CFL, the FL server collects the local model from the clients and then aggregates into the global model using the federated averaging algorithm (FedAvg) \cite{pmlr_v54_mcmahan17a}. Meanwhile, all states regarding the participant are collected in the FL server for selecting clients. On the contrary, in the DFL, the functions of the server are placed into all clients, including broadcast model, uploading model to other clients and aggregating model. Hence, the role of the server is removed from FL. Compared with CFL, DFL is suitable for dynamic networks and has scalability and robustness. However, in the DFL, wasting precious network resources exists because a model is transmitted multiple times among the clients.

\begin{figure}
    \centering
    \includegraphics[width=1.0\textwidth]{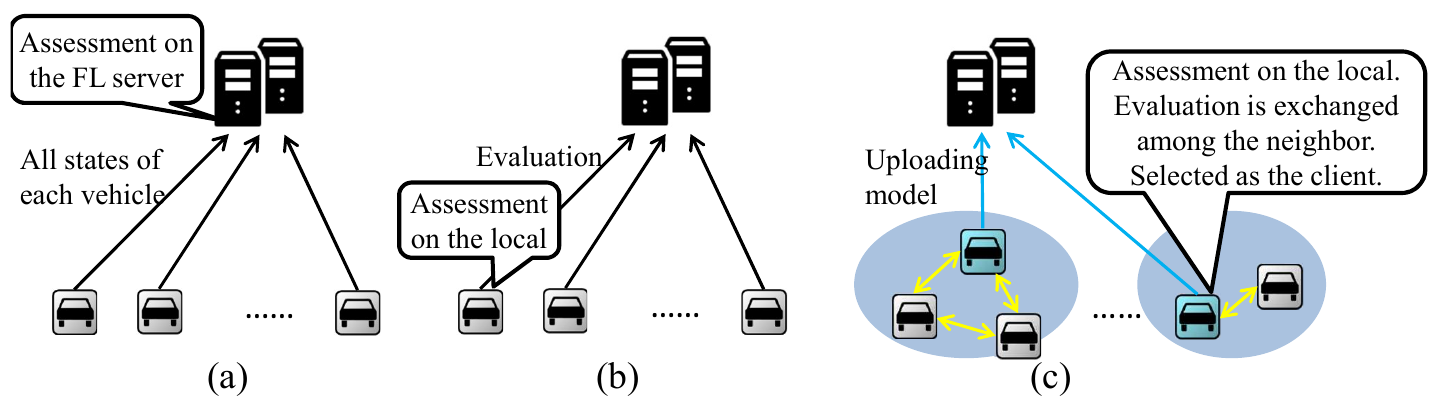}
    \caption{Different client selection schemes in FL. (a) Client selection in the CFL. (b) Client selection in the CFL-fuzzy \cite{fuzzyCentralized2022}. (c) Distributed client selection.}
    \label{clientselection}
\end{figure}

Following, we discuss three client selection schemes, as shown in Fig \ref{clientselection}, specifically, client selection in CFL, client selection in CFL-fuzzy \cite{fuzzyCentralized2022} and distributed client selection, respectively. In the CFL,
the whole states of each vehicle are collected to the FL server and then execute the following steps, assessment, sort, selection, broadcast global model to all clients, local training, uploading local model as well as aggregation, as shown in the Fig. \ref{clientselection}a. The FL server in the CFL plays the role of coordinator. For the client selection in the CFL-fuzzy, the assessment of each participant is processed locally and then updated to the FL server. Next, the following steps are run, sort, selection, broadcast global model to all clients, local training, uploading local model and aggregation, as shown in the Fig. \ref{clientselection}b. In the distributed client selection, the global model is broadcast to all participants at the start of each round, and the assessment is processed on the local. And then, the evaluation is exchanged among the neighbours and selects the client. Next, the selected client trains the model over on the local dataset and uploads their local model to the FL server. Finally, the FL server aggregates all local models received from the clients, as shown in the Fig. \ref{clientselection}c. The detailed process is presented in the section \ref{dcsframework}. The characteristic of distributed client selection is that selecting client adopts the distributed scheme as aggregating model adopts the centralized scheme. Furthermore, the FL server can not know which participants are selected as the client. The scheme can minimize the overhead caused by maintaining the active state of all participants while keeping the high efficiency of centralized aggregation. 

\subsection{Communication overhead}
\label{commOverHead}

In the FL system, the communication overhead comprises two parts, the overhead by exchanging models and the overhead by maintaining the active state of all participants. In the dynamic network (e.g. IoV), the participant's state needs to be constantly updated to the coordinator, such as the FL server, because the resources vary with time continuously and notice to the coordinator that a participant is still alive. The state changes may increase the chance of being selected as a client. So, all participants update the active state to be chosen as a client by the server. In general, the size of maintaining an active state is far larger than the size of exchanging model when millions of participants exist in the FL system. Following, we analyze a real example from Gboard \cite{gboard} to compare the two kinds of overhead. We choose transmitted data size as a comparing metric. Here, \textit{N} is the number of all participants, and $\tau$ represents the interval of sending state. \textit{s} represents the size of the state, which includes participant ID, resource information (e.g. computational capability, available network throughput, sample quantity and so on), and other information (e.g. vehicle position, acceleration, energy and so on). \textit{t} represents the length of a communication round. \textit{m} represents the size of the model. The transmitted data size for maintaining the active state of all participants is defined by Equation \ref{commOverhead}.

\begin{equation}
    c = \frac{N * s * t}{\tau}
    \label{commOverhead}
\end{equation}

The parameters referred from \cite{gboard} are listed in the table \ref{gboardpara}.

\begin{table}[h]
    \centering
    \caption{The parameters referred from Gboard \cite{gboard}.}
    \begin{tabular}{c|c}
    \hline
    \textbf{Parameters} & \textbf{Value}\\
    \hline
         The number of whole device & 1.5 million\\ 
         \hline
         The period of a communication round & 72 second \\
         \hline
         Model size & 1.4 Million Byte\\
         \hline
         The number of selected client in each round (average) & 300 \\
         \hline
         The size of active state (CFL) & 100 Byte\\
         \hline
         The size of active state (CFL-fuzzy) & 30 Byte\\
         \hline
    \end{tabular}
    
    \label{gboardpara}
\end{table}

\begin{figure}
    \centering
    \includegraphics[width=0.6\textwidth]{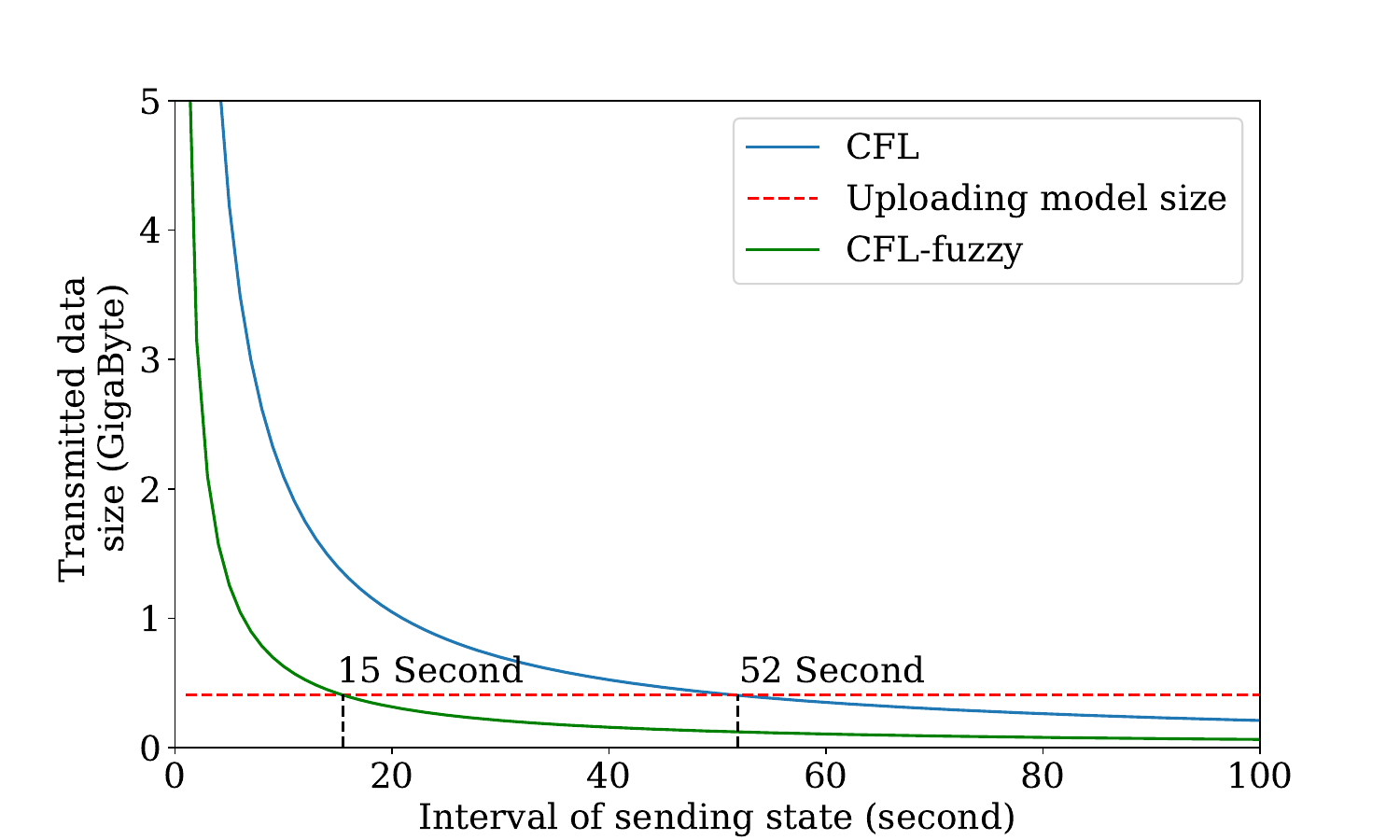}
    \caption{Comparison of two kinds of overhead. The dashed red line presents the size of the uploading model in each round. The blue and green lines present the overhead by maintaining the active state for all participants in the CFL and CFL-fuzzy, respectively.}
    \label{overhead}
\end{figure}

We compare with CFL and CFL-fuzzy \cite{fuzzyCentralized2022} in terms of the overhead maintaining active state for all participating devices in the Gboard. The dashed red line represents the size of the uploading model in each round over on the selected 300 client devices. The observation from Fig. \ref{overhead}, the size of overhead maintaining the active state of all participating devices reaches 22.5 Giga Byte in the interval of 1 second. In comparison, the uploading model size is only 0.41 Giga Byte. The size of maintaining the active state of all 
 decreases with the interval increase. Two curves, CFL and CFL-fuzzy, crossed with the uploading model size curve at 52 seconds and 15 seconds, respectively. However, in a dynamic network, such as IoV, some clients with poor performance are selected and dropped down the model convergence and even cause the traffic accident because the state of vehicles can not be updated in such an interval. The distributed client selection framework proposed can achieve low communication overhead and updating interval of the active state.

\subsection{Network model}

\begin{figure}
    \centering
    \includegraphics[width=0.6\textwidth]{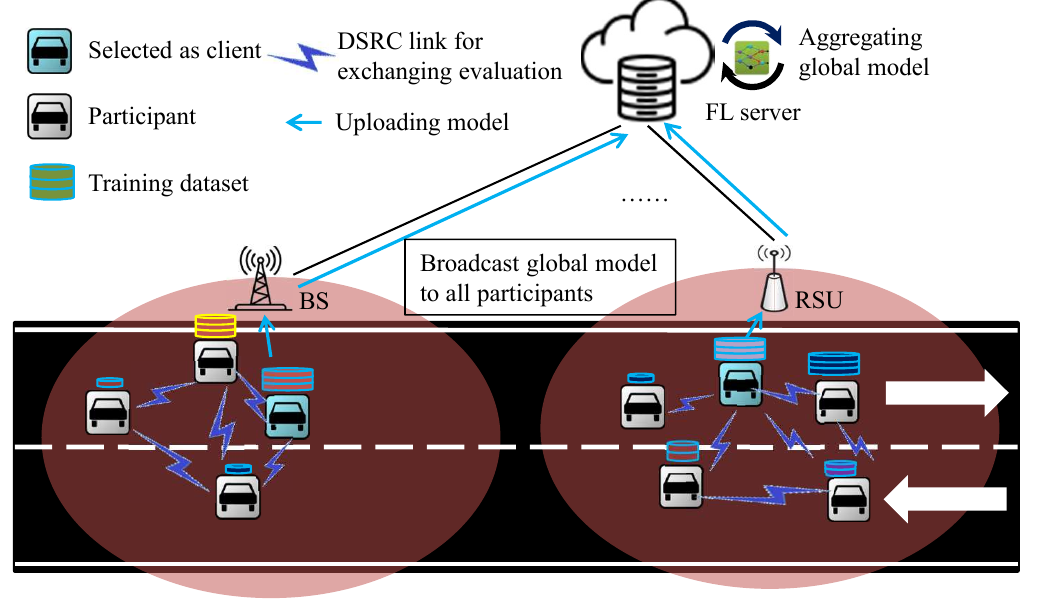}
    \caption{The network architecture of distributed client selection framework.}
    \label{system_model}
\end{figure}
We consider cellular-based IoV, in which every vehicle also supports dedicated short-range communication (DSRC) technology to exchange the evaluation between the neighbours, as shown in Fig. \ref{system_model}. Millions of participating vehicles are covered by multiple base stations (BSes) and share one FL server deployed in the cloud, to store a version of the global model in each round. The vehicle can move from one BS to another BS, and the throughput available in the network is affected by mobility. Each participating vehicle is willing to join the learning and share their knowledge with FL and other vehicles. Let $\Phi$ denote the set of participating vehicles and indexed by \textit{i}. Vehicle \textit{P}$_{i}\in\Phi$ owns some available computational resources as well as training samples \textit{D}$_{i}={(x_{i},y_{i})}$. Let \textit{C}$_{i}$ refer to the ratio of the computational capability in the vehicle \textit{P}$_{i}$ and $|\textit{D}_{i}|$ refers to the sample quantity in the vehicle \textit{P}$_{i}$, respectively. Each BS can schedule wireless resource blocks (RBs) to allocate the client for the transmitting model. An X2 interface links with neighbouring two BSes to support the handover efficiently. The wireless resource block allocation strategy is assumed independent and does not interfere with each other. BS allocates RBs through ``MAX C/I'' (Max Carrier to Interference) scheduling when multiple clients apply simultaneously.

\subsection{Distributed client selection framework}
\label{dcsframework}
We considered how much a client contributes to the global model in this framework. The client with a larger loss function has more contribution. So, the loss function of the dataset is introduced and used as one of the input variables in the multi-objective evaluation. To this end, the global model is broadcast to all participating vehicles at the start of each round. And every vehicle calculates the loss function without updating the model locally. The algorithm \ref{alg1} presents the whole process in each round.

\begin{algorithm}\footnotesize
\caption{The process of distributed client selection framework}

\label{alg1}
\begin{algorithmic}
\State \textbf{FL server:}
\State Broadcast global model to all participating vehicles.
\While {The deadline is not expired}
\State Receive the local models from the clients.
\EndWhile
\State Aggregate the local models using Eq. \ref{equ4_4} to generate global model for the next round.

\end{algorithmic}
\begin{algorithmic}
\State 
\State \textbf{Each participant $P_{i}$:}
\State Receive global model.
\State Calculate loss function on the local using following Equation

$l_{i} = \frac{\sum_{i=1}^{|D_{i}|} L_{i}(w_{i}^{k}, x_{i}, y_{i})}{|D_{i}|} $. \Comment{No updating model.}

\State Get evaluation $E_{i}$ using the multi-objective evaluator.
\If {$E_{i}$ $\geq$ $E_{\tau}$ } 

\Comment{$E_{\tau}$ is constant and used as a threshold.}
\State $E_{i}$ is broadcast to the neighbours.
\EndIf
\If {$E_{i}$ is the largest among the neighbours}
\State $P_{i}$ is a client.
\State Training model over the local dataset.
\State Uploading model to the FL server.
\EndIf

\end{algorithmic}
\end{algorithm}

Notably, the broadcasting model in the distributed client selection is the model that is transmitted to all participating vehicles. From a technical, reliable broadcast has yet to exist. Fortunately, the transmitting model need not require reliable transmission, and FL has a certain tolerance to the error of the model parameters in the transmission stage. Hence, the broadcast can implement by some technology like a multicast stream. 

\section{Multi-objective evaluator}
\label{section5}

In this section, we describe indispensable parts of the evaluator, including the prediction of the network throughput available and the time taken for the training. And then, four input variables for the fuzzy evaluator are presented. Next, the fuzzy evaluator, including the fuzzy rule, normalization to the input variable, and final evaluation, are explained. Finally, exchanging evaluation process is illustrated.

The evaluator is another important component of the distributed client selection, which is run in each participating vehicle. In the evaluator, we considered four variables, specifically, sample quantity, network throughput, computational capability, and loss function of the local dataset. Those variables can be obtained locally. Considering the non-existence of the closed-form solution over the four variables mentioned above, we adopt fuzzy logic as the evaluator, named fuzzy evaluator. A detailed description of the fuzzy evaluator is presented in Section \ref{fuzzy_evaluator}. 

\subsection{Prediction to the network throughput}
\label{Prediction_of_available_throughput}
The network environment can directly affect exchanging model between the FL server and the clients. In general, the network throughput at some time can be predicted according to the historical transmitting state in the past.

Communication between the FL server and the client mainly consists of two parts in each round, specifically, broadcasting the global model and uploading the local model to the FL server. The time to broadcast the global model does not affect the performance of the FL since the time can be considered as a constant in each round \cite{fedmax}, and the constant would not change anytime and anywhere. The time to upload the local model is the main component in FL communication. 

We consider reliable transmitting protocols, such as transmission control protocol (TCP), used as exchanging model protocol to upload the local model to the FL server. Therefore, TCP (Reno), a widely used protocol, is adopted to transmit the local model to ensure the trust and reliability of the model with the best effort.

The available throughput of the participating vehicles varies with the mobility of the vehicles. In practice, precisely predicting throughput is necessary for each participant when the fuzzy evaluator assesses the participating vehicle. To predict the throughput available, the sender's congestion control (CWND\_SND) window size in the TCP (RENO) is used to approximate the throughput of the participants. The assumptions are that every participating vehicle plays the sender's role in sending the data to the FL server, and the history record of CWND\_SND is stored in the sender when the data are transmitted. The available throughput of the participating vehicles achieves by averaging the CWND\_SND values within a certain period. 

The clarification is that the value of available throughput need not be exact, and obtained value meets some criteria that keep order relatively. In other words, the order of predicted throughput of the participating vehicles also keeps the order in terms of the real throughput in the real world. Since the evaluator only requires sorting the participating vehicles by the available throughput. In the real world, because of user privacy, collecting the information from both sender and receiver is impossible to predict the throughput. The congestion window size can reflect the variety of available throughput while the network environment changes with mobility.


\subsection{Training Time}
Because of the characteristic of heterogeneity, participating vehicles owning the computational capability differ from each other. Meanwhile, the training dataset distributes uniformly over participating vehicles hardly in terms of the sample quantity and classification. The time spent in the training is not identical because every participant has a different computational capability and Non-i.i.d dataset (Here, non-i.i.d refers to the feature of independent and identically distributed). The drawback of the simulator is the time taken in training, which the client needs to learn previously. Moreover, the simulation process must appear the heterogeneity of the FL system mentioned above. Hence, the time taken in the training is calculated by the following Equation \ref{eq1}. 

\begin{equation}
    \textit{T}_{\textit{i}}^{\textit{com}}=\frac{EC_{i}|\textit{D}_{i}|}{B_{size}B_{exe}},
    \label{eq1}
\end{equation}
where \textit{B$_{size}$} refers to the batch size and \textit{E} refers to the number of epochs in the learning. \textit{B$_{size}$} and \textit{E} are described as a constant and are the same for all participants in the FL learning process. \textit{B}$_{exe}$ is denoted the time to train the model on the client for \textit{B}$_{size}$ samples. The value of \textit{B}$_{exe}$ averages real value, which is obtained from a huge amount of the experiments conducted on PyTorch \cite{paszke2019pytorch}. Conducting experiments on the environments is described as follows. The hardware and software configurations are Intel@Core™ i5 multi-core processor, CPU@2.50GHz×8 core, RAM@16GiB, and PyTorch@1.8 version without GPU. 

\subsection{Fuzzy evaluator with multi-objective}
\label{fuzzy_evaluator}

Fuzzy logic is an approach that does not need a close-form solution over considering the variables and can obtain the list of the output values. Having the characteristic of the lightweight, fuzzy logic run on the participating vehicles. We considered four input variables, specifically, sample quantity, throughput available, computational capability, and loss function of the dataset, which are related to the uploading model success rate as well as the contribution of the global model. These input variables are essential to evaluate whether a participating vehicle is ``good'' or ``bad'' for the FL. The reasons are listed as follows. On the one hand, the distance between the vehicle and BS/RSU varies with the time domain, and the throughput also fluctuates. Similarly, the computational capability is also frequently changed over time. The two input variables above are the main factor affecting the uploading model's success rate. On the other hand, the dataset with more samples contributes more to the convergence. Meanwhile, in the FL with the non-i.i.d feature, the diversity of the dataset across the participating vehicles can accelerate the convergence and be measured by the loss function. The greater the loss function, the more the diversity of dataset \cite{pmlr-v151-jee-cho22a}. Therefore, the sample quantity and the loss function are introduced to measure the quality of the dataset. Following, we present the input variables and its description.

\begin{figure}
    \centering
    \includegraphics[width=0.8\textwidth]{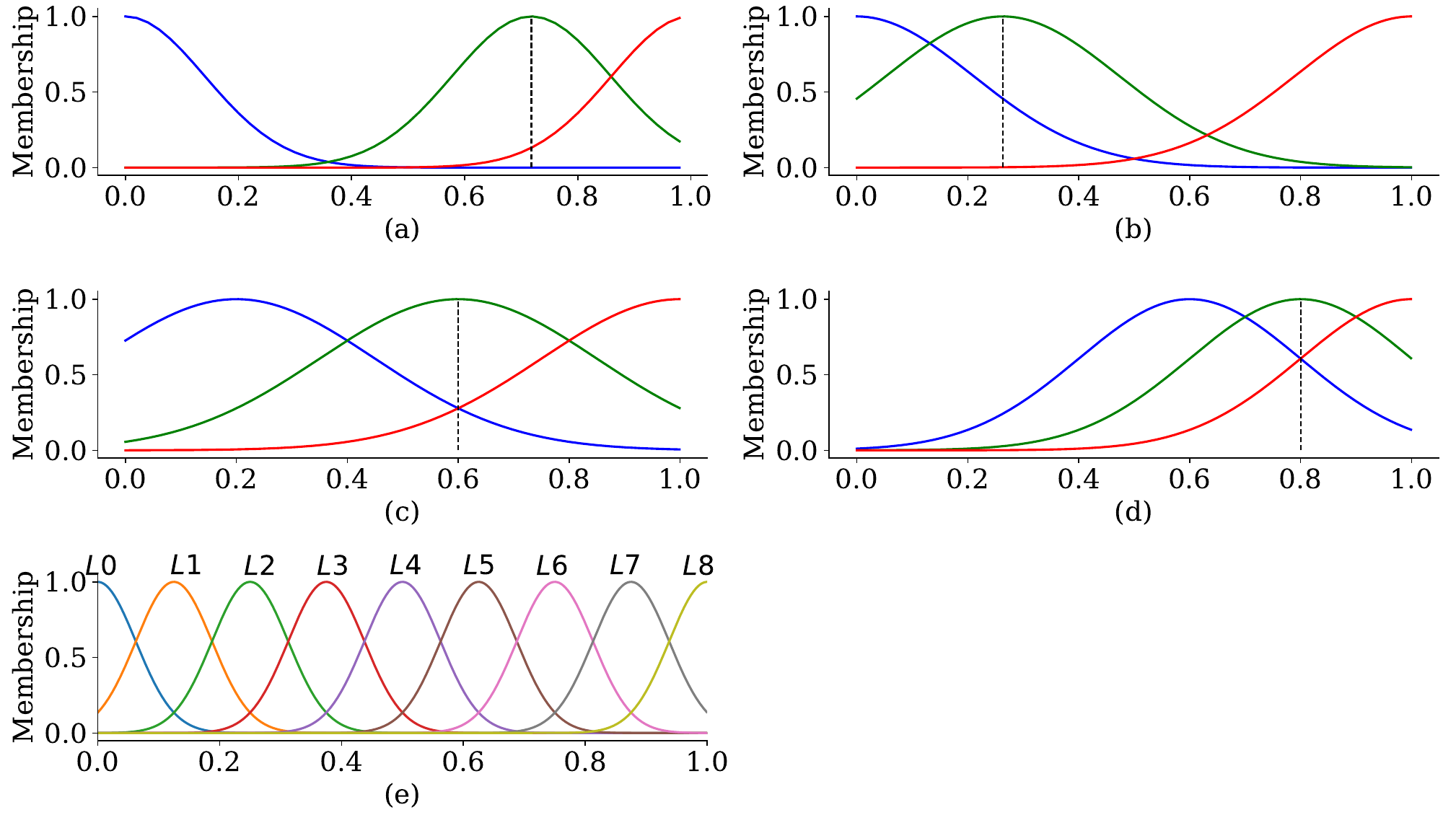}
    \caption{Membership functions used in the fuzzy evaluator. (a) Normalized sample quantity. (b) Normalized available network throughput. (c) Normalized computational capability on the local. (d) Normalized loss function training on the local dataset. (e) Evaluation mapped into different levels. In subfigure (a) - (d), the red line represents that the participant has better performance (or owns more resources) on a specific factor. The green/blue lines mean that the participant has average/poor performance (or owns average/less resource) on the specific factor. The dashed line in subfigure (a) - (d) represents the mean value of each variable.}
    \label{membership}
\end{figure}

\textbf{Sample quantity (SQ)}: The convergence can be accelerated when more samples are trained in machine learning. Similarly to FL, the client with more samples participates in the FL, speeding up the convergence. Hence, the participant with more samples should be selected as the client to join the FL. Considering the number of clients, selecting as many clients as possible is equivalent to training more samples in the round. However, the number of clients must be restricted because of the bandwidth limitation. Selecting a client with more samples is more efficient than the method that selects many clients. Therefore, the fuzzy evaluator uses the sample quantity as an input variable. The value of the sample quantity is normalized into [0, 1]. And then, the normalization is mapped into three levels, sufficient, average, and shortage, as shown in Fig. \ref{membership}a.

\textbf{Throughput available (TA)}: The throughput determines whether the model is uploaded successfully. This variable represents the network environment of the vehicle and is affected by the number of nodes around, allocated RBs, and the distance to the BS/RSU. Obtaining throughput available is described in Section \ref{Prediction_of_available_throughput}. Fig. \ref{membership}b shows the membership function of throughput available. Similarly to the sample quantity, the throughput available is normalized and mapped into three levels, good, middle, and poor, respectively, as shown in Fig. \ref{membership}b.

\textbf{Computational capability (CC)}: The computational capability determines how fast the learning is. This variable denotes the available computing power of the participant, which is one of the factors affecting training time. Similar to the sample quantity, the computational capability is also normalized and mapped into three levels, strong, middle, and weak, respectively, as shown in Fig. \ref{membership}c.

\textbf{Loss function (LF)}: Training various samples can improve the generalization ability of the model. Selecting a client with diversity has more contribution to the convergence. In the paper, we adopt the loss function to measure the diversity of the dataset and calculate by Equation \ref{lossfunction}. 

\begin{equation}
    \centering
    \label{lossfunction}
    l_{i} = \frac{\sum_{i=1}^{|D_{i}|} L_{i}(w_{i}^{k}, x_{i}, y_{i})}{|D_{i}|} 
\end{equation}

The loss function calculation is the same as the loss function in the training but not updating the gradient. In addition, all samples need to calculate the loss function once to average the error. The shuffling sample does not affect the result without updating the model. The greater loss function represents that the dataset owns higher diversity and some new features. The loss function is also normalized and mapped into three levels, greater, middle, and lower, respectively, as shown in Fig. \ref{membership}d.

All variables adopt the Gaussian function as the membership function to ensure that different input value results in different output for the evaluation. The dashed line represents the mean of the input variable calculated from the historical records.

For the unity of the expression, all input variables are normalized into [0, 1] by the Equation (\ref{eq2}). In Equation (\ref{eq2}), two variables named ``$Value$" and ``$Maximum \quad of \quad input \quad variable$" need to be replaced by the actual value of a specific input variable.

\begin{equation}
    Normalized = \frac{Value}{Maximum \quad of \quad input \quad variable} \times 100\%.
    \label{eq2}
\end{equation} 


The fuzzy evaluator comprises four components: fuzzification, fuzzy rules, defuzzification, and client selection. Mamdani Method is used as the fuzzy inference technique. Regarding the fuzzification of the output value, the output of four variables is mapped into three levels and as shown in Fig. \ref{membership}a - Fig. \ref{membership}d. And then, the normalization of the input variable is associated with the output variable through fuzzy rules and output to nine levels from $L_{0}$ to $L_{8}$. Detailed fuzzy rules are listed in Table \ref{fuzzyrule}. 

Fuzzification is that the crisp input value needs to be transformed into three different linguistics. These linguistics have been described in Section \ref{fuzzy_evaluator}. Moreover, the bound of each linguistic is defined through historical records.

The fuzzy rule contains 81 items since there are four input variables, and each input variable is mapped into three linguistics, as shown in Table \ref{fuzzyrule}. Each item in Table \ref{fuzzyrule} is implemented by a simple IF-THEN logic with single or multiple antecedents. Finally, all antecedents are outputted to one consequent. Those rules are essential to the evaluation of participating vehicles. Many experiments are conducted to decide the mapping relationship between the input and output, and the experiment with the best performance is selected as the item of the fuzzy rule. 

Defuzzification is that the output needs to be transformed to a scalar through the centre of gravity (COG), one of the most commonly used methods. COG is defined as shown in Equation \ref{eq_cog}.
\begin{equation}
    y^{*} = \frac{\sum_{i=1}^{n}{a_{i} \cdot \mu({a_{i}})}}{\sum_{i=1}^{n}\mu({a_{i}})}
    \label{eq_cog}
\end{equation}
where $a_{i}$, $\mu({a_{i}})$, and \textit{n} are denoted the sample element, the membership function, and the number of the element in the group, respectively. Fig. \ref{eq_cog} illstrates COG. The value of 58.09 in Fig. \ref{eq_cog} represents an output calculated by Equation \ref{eq_cog}, and the value belongs to the L6 level.

\begin{figure}
\centerline{\includegraphics[width=0.6\textwidth]{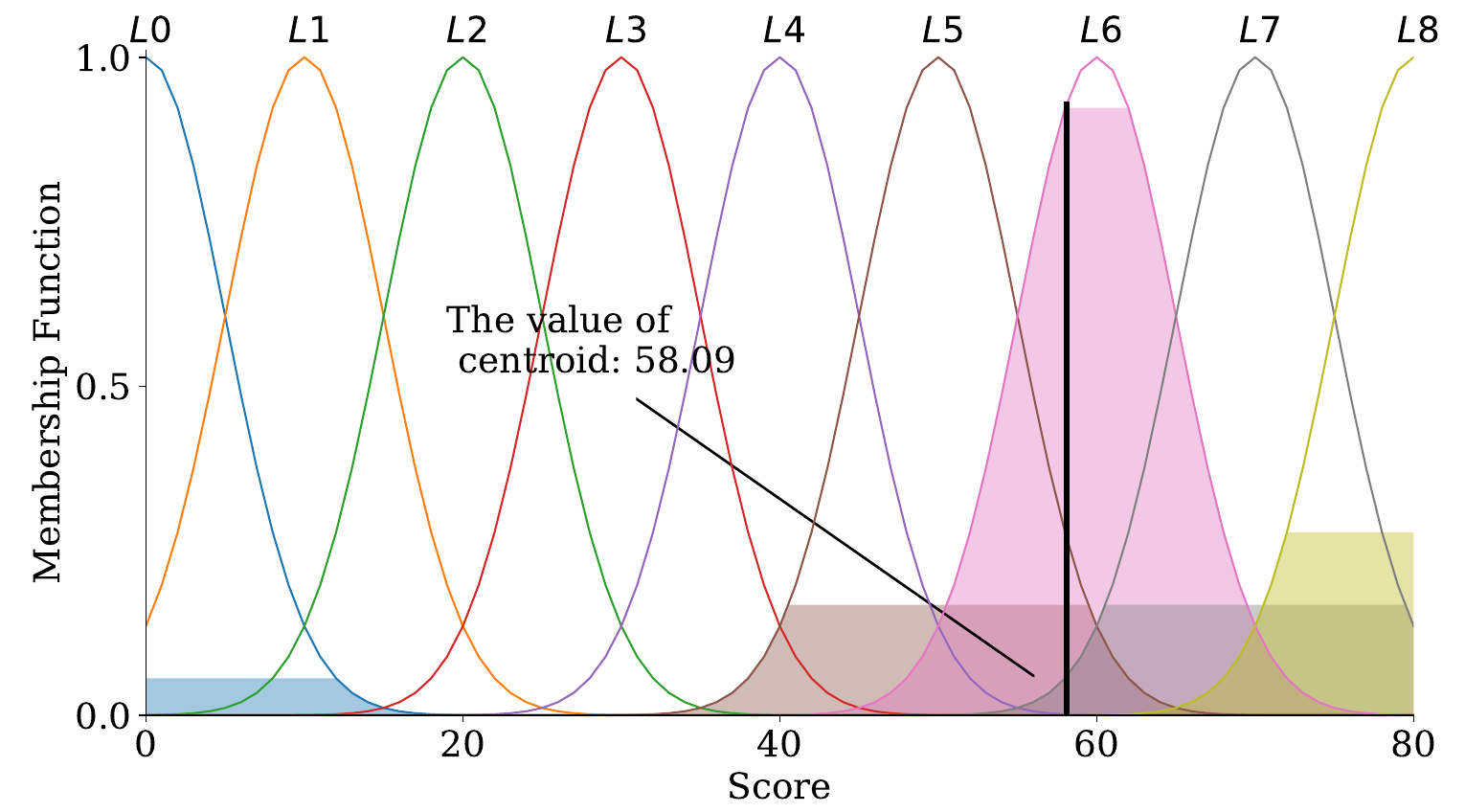}}
\caption{The center of gravity (COG). }
\label{cog}
\end{figure}

The final evaluation is broadcast over DSRC communication to all neighbours. And each participating vehicle maintains a table to store and update the evaluation received from the neighbours and itself. Ultimately, checking the table's top \textit{m} contains its id. The participating vehicle becomes a client while the table contains its id and vice versa. Following this, the selected client trains the dataset and uploads their local model. 

\begin{table}
\caption{Parts of fuzzy rule}
\label{fuzzyrule}

\begin{tabular}{ c|c|c|c|c|c }

\hline
 & \textbf{SQ} & \textbf{TA} & \textbf{CC} & \textbf{LF} & \textbf{Evaluation} \\
 \hline
 1 &  Sufficient & High & Strong & Greater & L8\\
\hline
2 &  Average & High & Strong & Greater & L7\\
\hline
3  &  Shortage & High & Strong & Greater & L6 \\
\hline

... & ... & ... & ... & ... & ...\\
\hline
52 &  Sufficient & Poor & Weak & Middle & L2\\
\hline
53 &  Average & Poor & Weak & Middle & L1\\
\hline
54  &  Shortage & Poor & Weak & Middle & L0\\
\hline

... & ... & ... & ... & ... & ...\\
\hline
79 &  Sufficient & Poor & Weak & Smaller  & L0\\
\hline
80  &  Average & Poor & Weak & Smaller  & L0 \\
\hline 
81  &  Shortage & Poor & Weak & Smaller  & L0 \\
\hline
\multicolumn{6}{c}{\textit{* SQ: Sample Quantity, TA: Throughput Available, }}\\
\multicolumn{6}{c}{\textit{CC: Computational Capability, LF: Loss Function.}}\\
\end{tabular}
\end{table}

\section{Simulation and evaluation}
\label{section6}
\subsection{Set up}

  A realistic wireless vehicular network simulator is integrated with OMNeT \cite{omnet2010}, simuLTE \cite{simulte2014}, SUMO \cite{SUMO2018}, and Pytorch \cite{paszke2019pytorch}. The number of vehicles is 30, and all vehicles are running on a straight road with 1000 m in length and follow the free-way model. Each vehicle has two communication interfaces: cellular and DSRC (like the IEEE 802.11p interface). In the network, multiple base stations are located uniformly on the map, allocating wireless resources to the vehicle. Wireless resources are allocated to up/downlink streams identically. The available throughput of the participating vehicle reaches 10.4 Mbps when enjoying the highest modulation coding scheme (MCS) and the whole wireless resources. On the contrary, the available throughput only reaches 0.24 Mbps under the lowest MCS and the whole wireless resources. The evaluation of the participating vehicle is broadcast to the neighbouring vehicles over the IEEE 802.11p interface in a fixed interval. Each participating vehicle needs to maintain a table which stores the evaluation of the neighbour in the specific range. Furthermore, the content of the table is also updated in fixed intervals. To avoid the occurrence of stragglers, the deadline of the communication round is introduced in the simulator and set to 20 seconds. The local model received after the deadline is discarded.

\begin{table}[h]
    \caption{Configuration of the simulator for distributed client selection.}
    \label{configuration_distributed}
    \centering
    \begin{tabular}{p{7cm}|p{5cm}}
    \hline
    \textbf{Parameters} & \textbf{Value}\\
    \hline
    RBs of upstream/downstream & 1:1 \\
    \hline
    Batch size & 20 sample\\
    \hline 
    Epochs & 30 \\
    \hline
    Execution time of one batch $B_{exe}$  & 0.06 s \\
    \hline
    Highest throughput (cellular network) & 10.4 \textit{Mbps} \\
    \hline
    Worst throughput (cellular network) & 0.24 \textit{Mbps} \\
    \hline
    Range of exchanging evaluation over DSRC & 200 meters\\
    \hline 
    The number of vehicles & 30 \\
    \hline
    Deadline of communication round & 20 second\\
    \hline 
    The number of selected clients in each area (distributed) & 2 \\
    \hline
    The length of road & 1000 meters; straight road \\
    \hline
    The location of the FL server & At 520 meters of the road (very close to the BS)\\
    \hline 
    The sample quantity of the vehicle & VehID 0-11: about 4500 images,
    VehID 12-29: about 45 images\\
    \hline 
    The vehicle distribution  & uniform\\
    \hline
    The packet size & 1500 byte\\
    \hline
    The latency from vehicle to cloud & 200 ms\\
    \hline
    The latency from vehicle to vehicle (DSRC) & 40 ms \\
    \hline
         
    \end{tabular}
    
\end{table}

A dataset regarding image recognition, MNIST \cite{deng2012mnist}, is adopted as the training dataset. To meet the characteristic of non-i.i.d, we synthesize the dataset with the non-i.i.d feature and the whole sample in MNIST is re-distributed over the participating vehicles according to the following rules. The local dataset of each vehicle can come from multiple classes which own identical quantity samples. These vehicles have an unbalanced dataset. For example, the vehicle's id numbered from 0 to 11 have about 4500 samples, while other vehicle's id numbered from 12 to 29 are only owning about 45 samples. All samples in the participating vehicles are not duplicated each other. Considering the vehicle density running on the road, the number of selected clients in each range of 200 m is up to 2. In the centralized client selection, the FL server selects 5 clients in each round. Parameters used in the simulator are listed in Table \ref{configuration_distributed}.

The learning model used in the FL has 7 layers, including two layers of convolution layer, one layer of flattened layer, two layers of max pooling layer, and two layers of the fully connected layer, to train the MNIST dataset. Each sample has a size of 28$\times$28 and single channels. The total number of the trainable variables in the learning model is about 1.66 million and takes the disk space to 5.2 Mbytes. All models are not compressed in the broadcasting and uploading stage.

\subsection{Evaluation}
In this subsection, we evaluate the performance compared to the baselines and the proposal in terms of the model accuracy, the influence on the distribution of the vehicle, the convergence over the non-i.i.d dataset as well as accumulated consumed time on the communication overhead.

To evaluate the performance of the proposal, we consider multiple benchmarks, specifically, centralized client selection (CCS) and centralized client selection with fuzzy logic (CCS-fuzzy) \cite{fuzzyCentralized2022}. CCS means that all information involving the active states of the client needs to be transmitted to the FL server, and the FL server is in charge of the client selection. Random client selection is a typical CCS scheme. CCS-fuzzy refers that the fuzzy evaluation to assess the client is moved from the FL server to the participating vehicles and uploaded to the FL server after evaluation. DCS refers to selecting a client which does not rely on the FL server. The evaluation of the participant is exchanged among the neighbours over DSRC communication. And then, these nodes elect some nodes with the highest evaluation to be acted as the clients for uploading the model. 

In the CCS-fuzzy and random scheme, the FL server selects 5 clients randomly from all vehicles. Some clients may become the stragglers in the selected clients. In general, the straggler can not upload their local model before the deadline expires because their computational capability and network throughput are too low.

\begin{figure}
    \centering
    \includegraphics[width=0.6\textwidth]{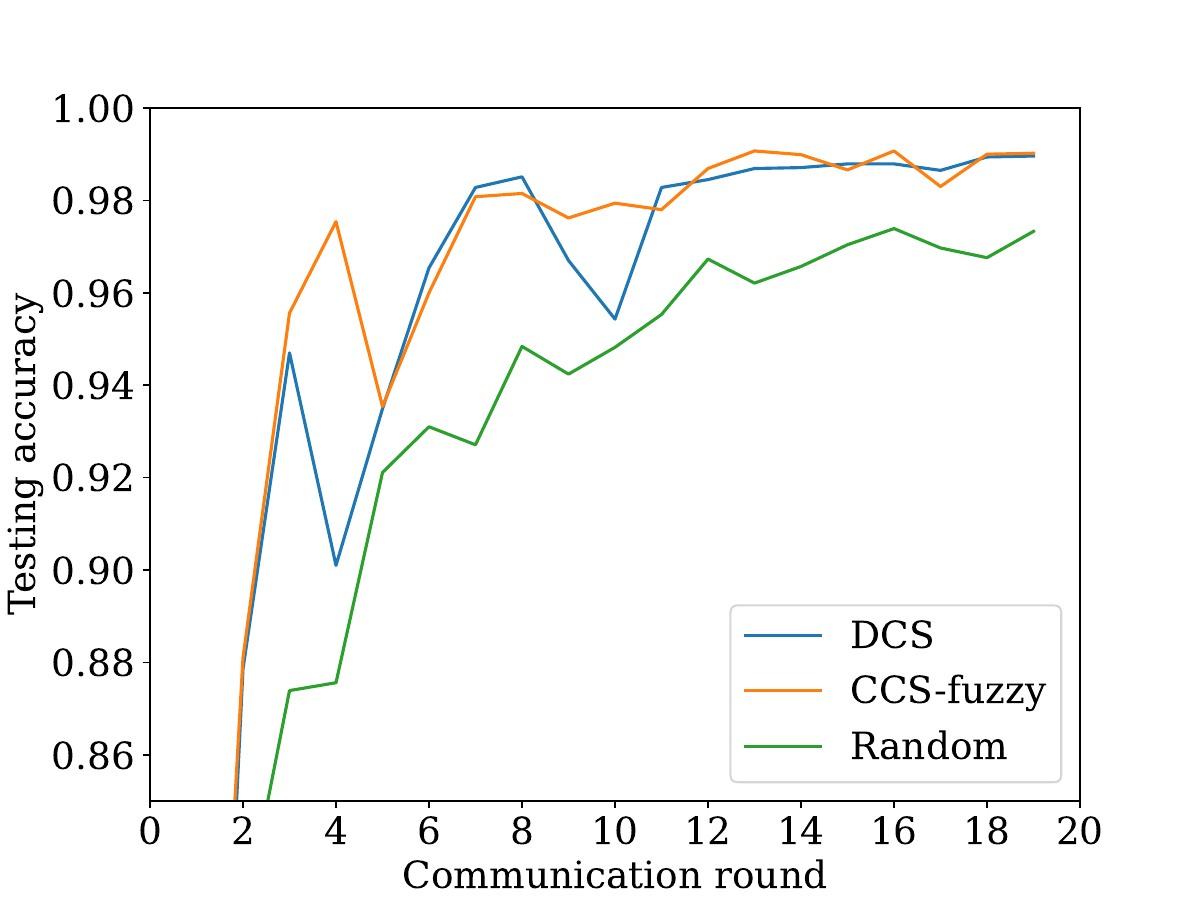}
    \caption{Accuracy on DCS, CCS-fuzzy and random scheme.}
    
    \label{random_distributed_centralized}
\end{figure}

Fig. \ref{random_distributed_centralized} compares the accuracy of DCS, random scheme and CCS-fuzzy. The sample quantity of the vehicle is the same as in Table \ref{configuration_distributed}. Every vehicle owns 9 classes, each containing an identical sample quantity. The number of selected clients in the DCS is averaged at 5.15. The number of selected clients in a random scheme and CCS-fuzzy is set to 5 as a constant. The results can be observed from Fig. \ref{random_distributed_centralized} that the CCS-fuzzy outperforms DCS and random scheme. The reason is that CCS-fuzzy can select the participating vehicle with the highest evaluation as the client and can accelerate the convergence. In other words, CCS-fuzzy can choose clients with better resources in terms of computational, communication and local datasets. However, it is notable that the proposal also performs well. The two curves of CCS-fuzzy and the proposed scheme overlapped in the final stage, while DCS largely jitters in the initial step. In conclusion, DCS can achieve the same level as CCS-fuzzy. Furthermore, DCS can outperform the random scheme after a certain communication round.

\begin{figure}
\centerline{\includegraphics[width=0.6\textwidth]{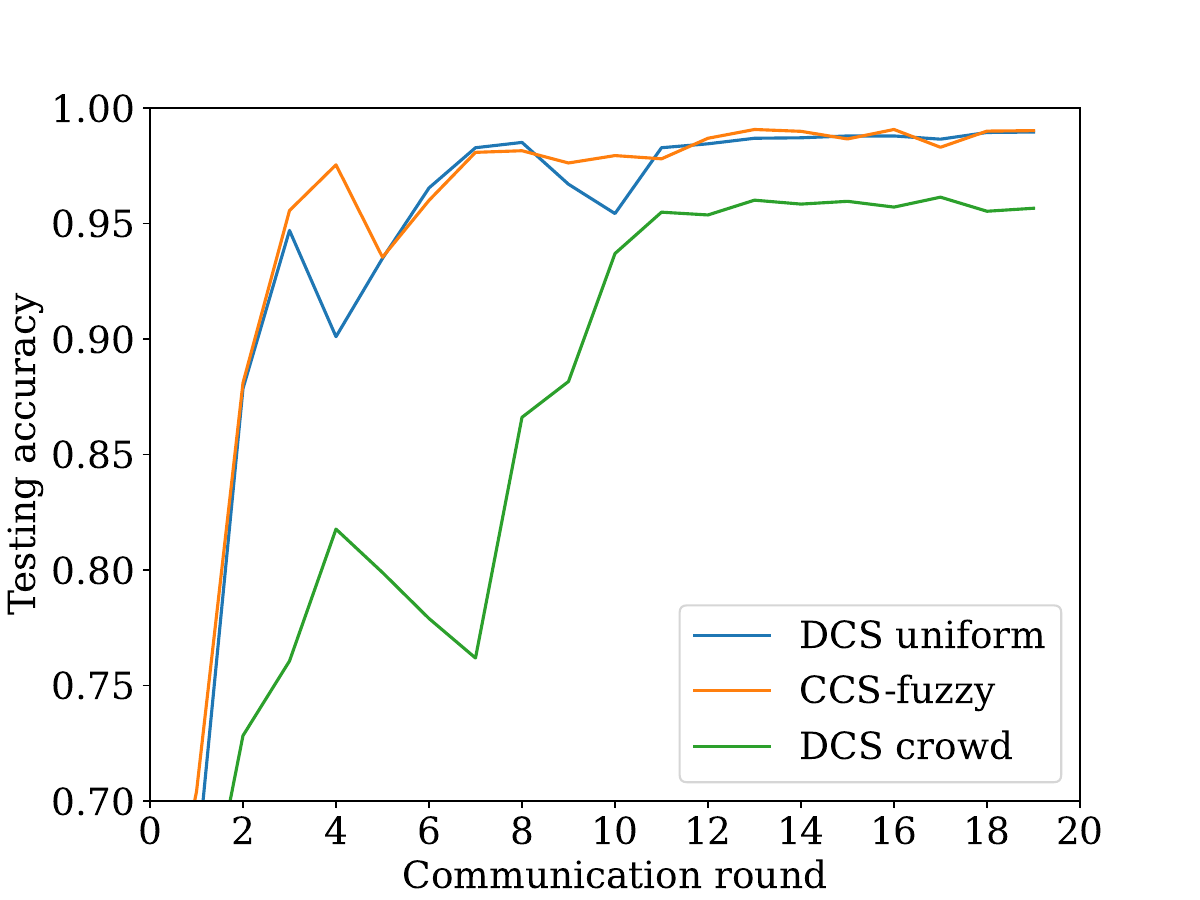}}
\caption{Different vehicle distributions influence the accuracy.}
\label{scenario1.2}
\end{figure}

The distribution of participating vehicles can impact the performance of DCS because DCS can select the optimal client in the neighbouring small area. To illustrate the case, we design two distributions for participating vehicles, including uniform distribution and extreme distribution, respectively. In uniform distribution, all vehicles are distributed randomly. In the extreme distribution, the vehicles with better evaluation are crowded in one small area, while the remaining vehicles with poor evaluation are crowded in another small area. In Fig. \ref{scenario1.2}, the results are illustrated the performance involving CCS-fuzzy, DCS with uniform distribution and DCS with extreme distribution. Other parameters are the same as in Table \ref{configuration_distributed}. The number of selected clients in the uniform distribution is averaged at 5.05, while the number of selected clients in the extreme distribution is 6 as a constant. The observation from Fig. \ref{scenario1.2} is that the accuracy of the uniform distribution is approaching the CCS-fuzzy scheme and is better than the performance of the extreme distribution. The reason is described as follows. The vehicles with better evaluation are more likely to be selected as the client in the uniform distribution. On the contract, the vehicles with better evaluation are crowded in the small, leading to ``cut-throat competition'' in the extreme distribution. Hence, a small number of vehicles with better evaluation are selected as the client and pulled down the convergence speed.

\begin{figure}
\centering
	\begin{minipage}[t]{0.3\linewidth}
		\centering
		\subfloat[Each client owns 9 of 10 classes.]{\includegraphics[width=1.0\textwidth]{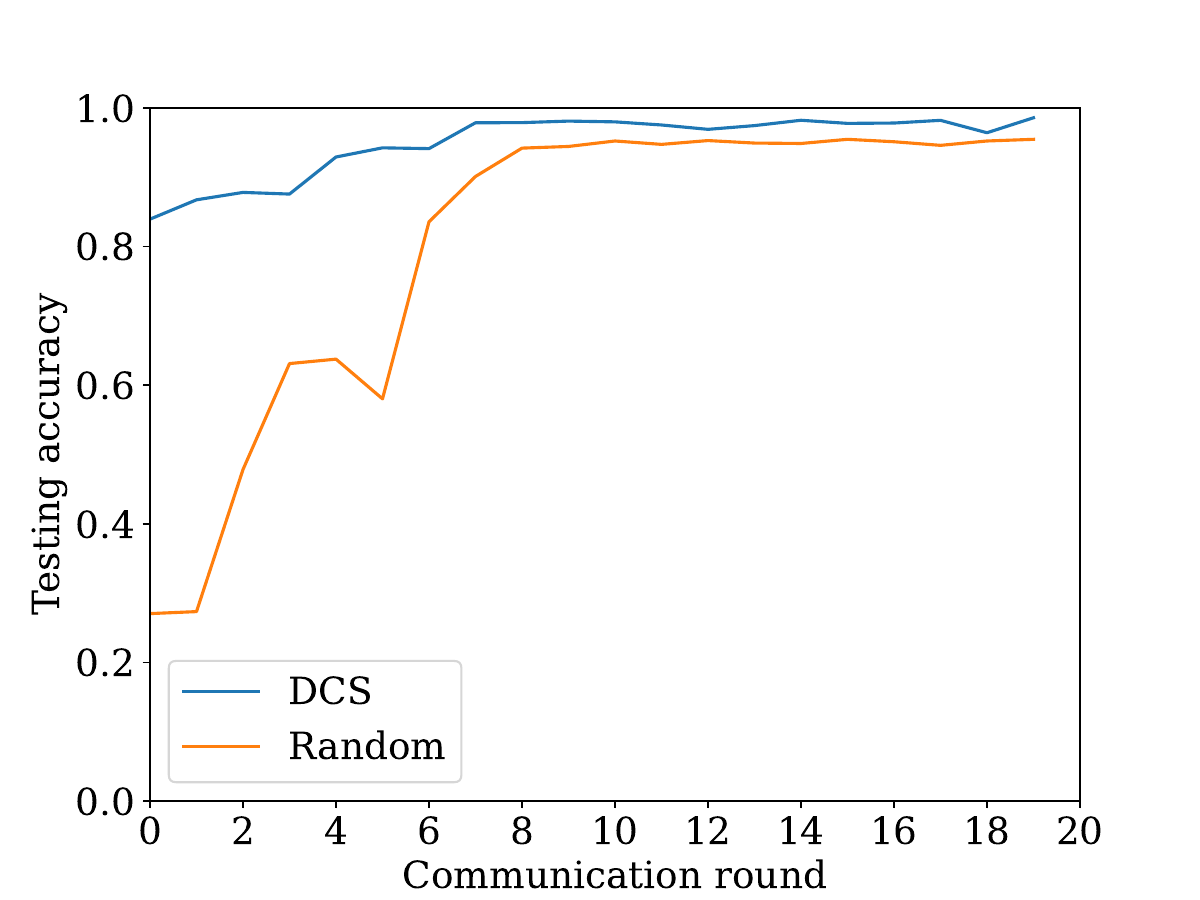}}
	\end{minipage}
	\begin{minipage}[t]{0.3\linewidth}
		\centering
		\subfloat[Each client owns 6 of 10 classes.]{\includegraphics[width=1.0\textwidth]{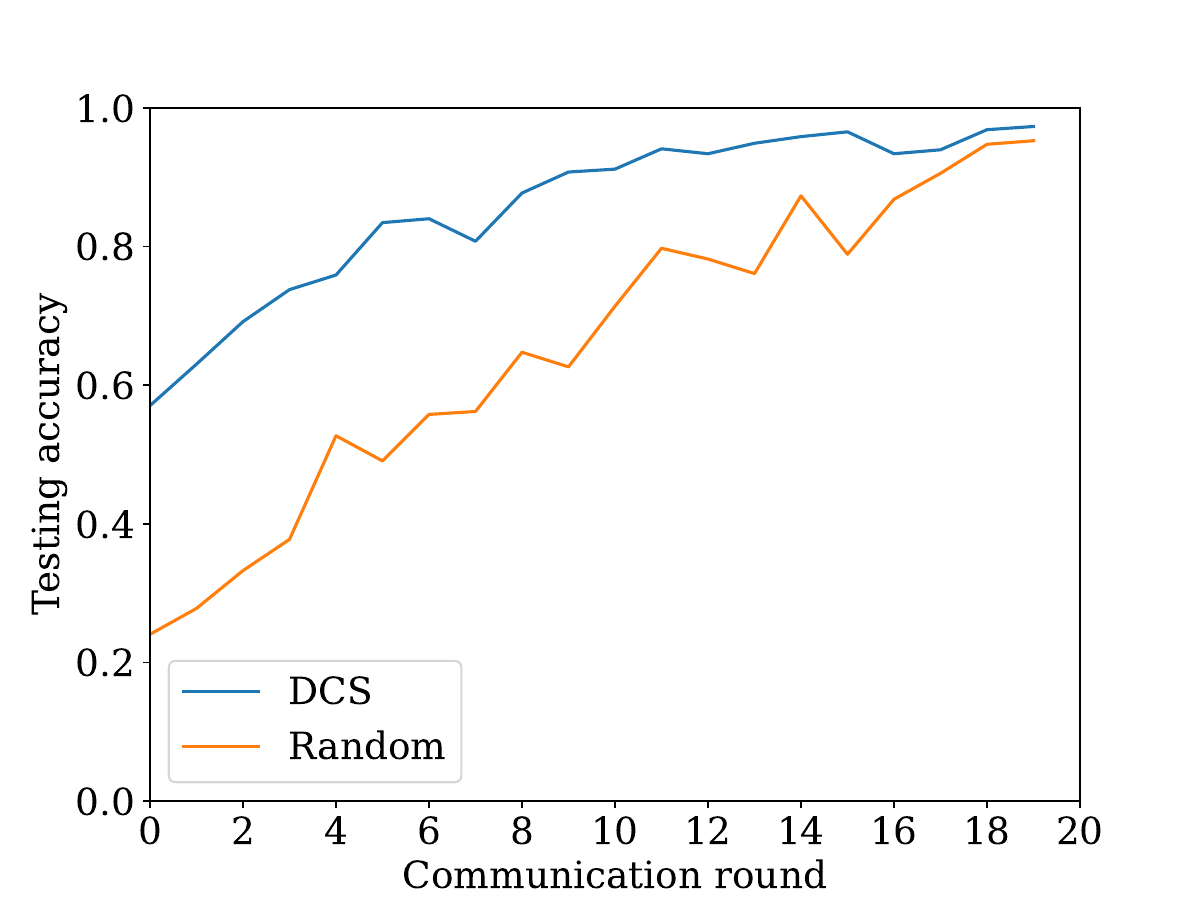}}
	\end{minipage}
	\begin{minipage}[t]{0.3\linewidth}
		\centering
		\subfloat[Each client owns 2 of 10 classes.]{\includegraphics[width=1.0\textwidth]{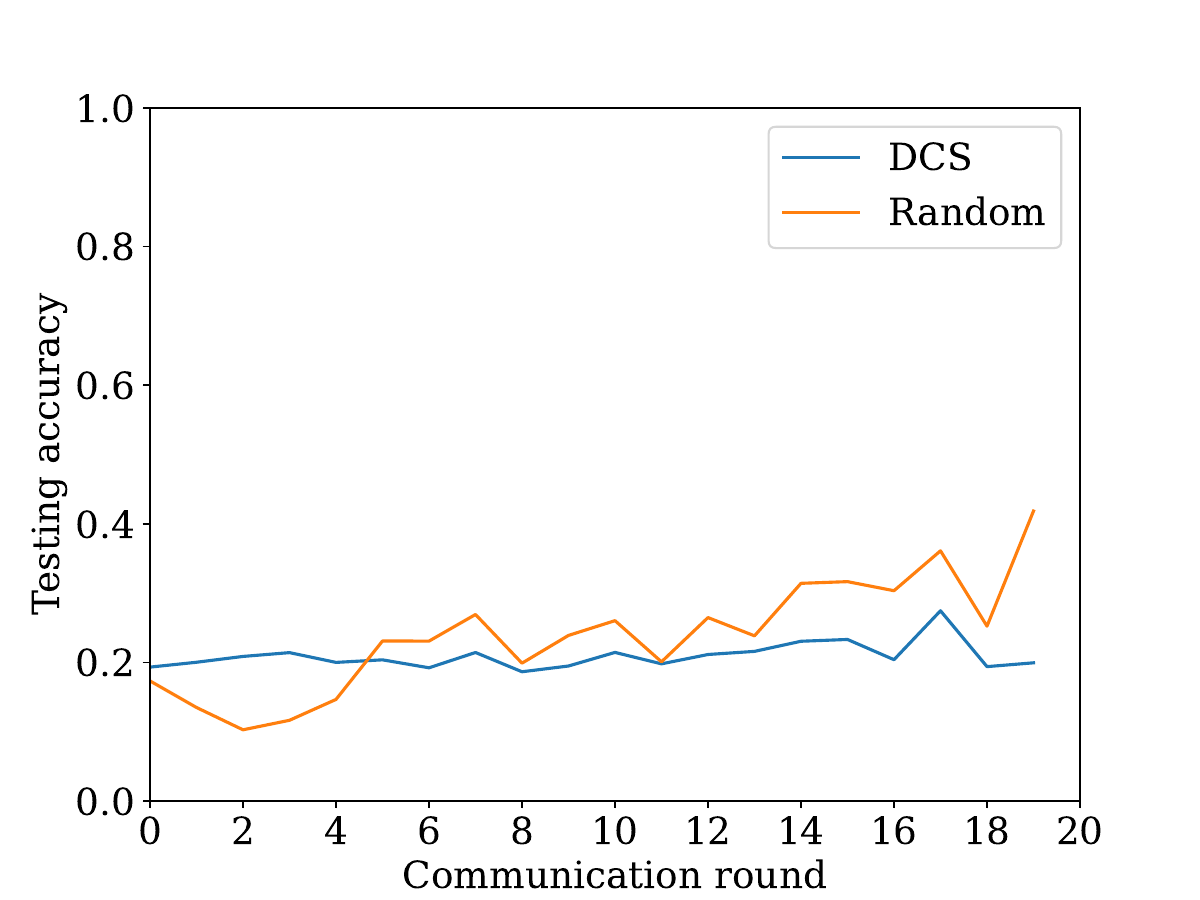}}
	\end{minipage}
\caption{Non-i.i.d characteristic impacts on the accuracy.}
\label{noniid}
\end{figure}


To show the performance regarding the non-i.i.d dataset, we run three experiments with an unbalanced quantity dataset, in which each vehicle contains 9 classes, 6 classes and 2 classes of 10 classes, respectively. Fig. \ref{noniid} compares DCS and random scheme over the non-i.i.d dataset. In Fig. \ref{noniid}a, Fig.\ref{noniid}b and Fig. \ref{noniid}c, the number of selected clients in DCS is averaged at 5.15, 5.2 and 4.95, respectively. The number of selected clients for the random scheme is set to 5 as a constant. The conclusion from Fig. \ref{noniid}a, Fig. \ref{noniid}b and Fig. \ref{noniid}c observed that the non-i.i.d characteristic of the dataset has a great impact on the model accuracy and the convergence speed. The convergence speed accelerates when the characteristics of the dataset are approaching from non-i.i.d to i.i.d. Because the non-i.i.d dataset increases the weight shifting in training. In the special case, such as without the intersection between the datasets, the accuracy can not meet the requirements and even can not converge, as shown in Fig. \ref{noniid}c. On the other hand, the proposal performs well compared to the random scheme but extreme non-i.i.d. The reason is that the loss function of local data is considered in the client selection stage. It makes to enhance the diversity of the dataset and decreases non-i.i.d characteristics.   

\begin{figure}
    \centering
    \includegraphics[width=0.6\textwidth]{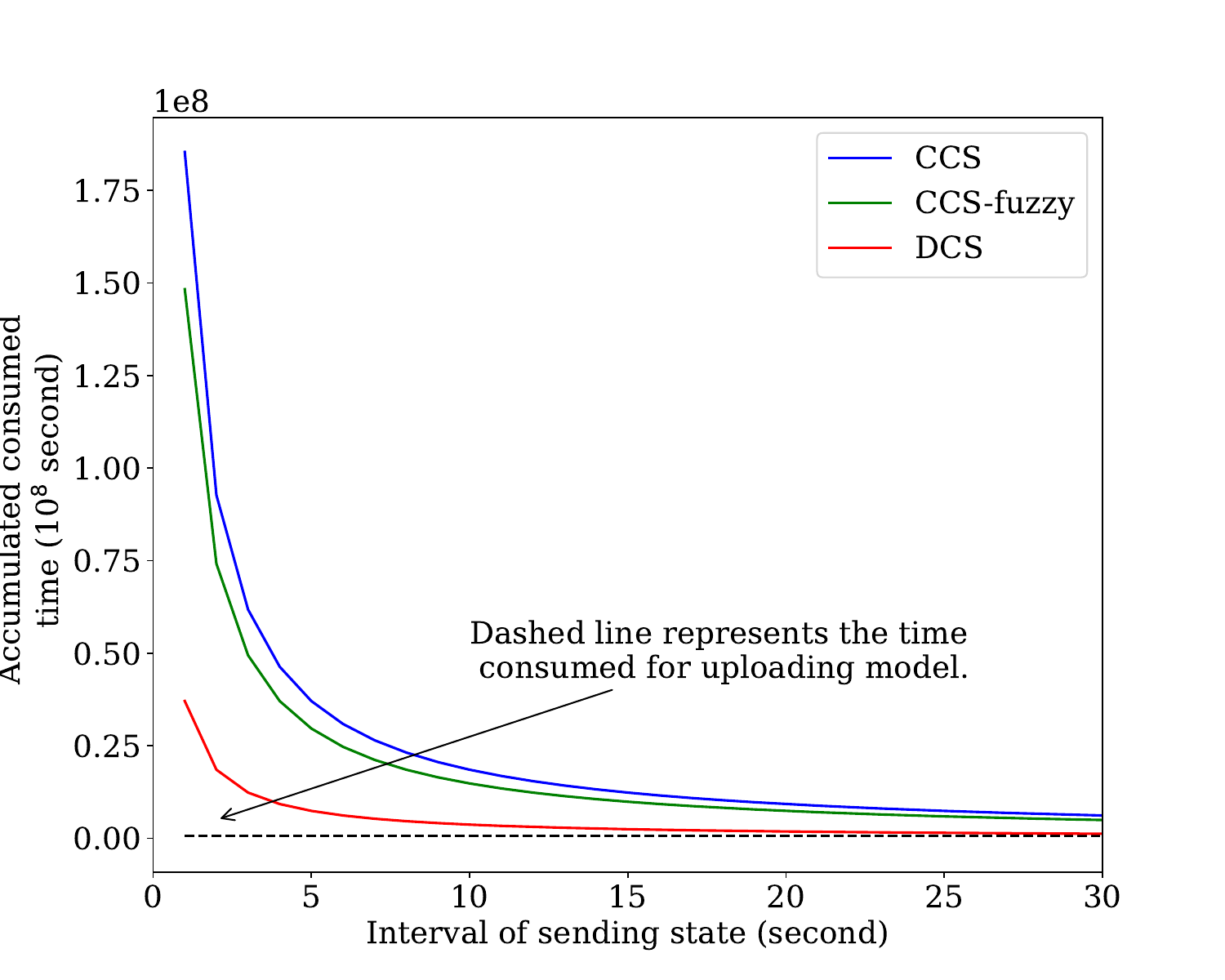}
    \caption{Accumulated consumed time vs sending interval.}
    \label{Accumulated}
\end{figure}

Finally, we analyze the time consumed on the communication overhead. We adopt the accumulated consumed time as a metric, the sum of time each participant consumed on the communication. We adopt Tokyo region as an example to analyze the communication overhead. According to the statistics \cite{numberCarTokyo}, by 2021, the number of registered motor vehicles reached 3.09 million in Tokyo region, Japan. In each communication round, 1, 000 vehicles are elected as the client. We consider that the sum of the total time consumed in communication, including exchanging the model and maintaining the active state of the vehicles. Sending an active state also spends a full latency since it is considered a small packet. And other parameters are listed in Table \ref{configuration_distributed}. Fig. \ref{Accumulated} compares the accumulated consumed time over DCS, CCS-fuzzy, CCS and exchanging the model. The dashed line represents the consumed time for exchanging the model. The following results can be seen in Fig. \ref{Accumulated}. First, compared to exchanging model, the cost caused by maintaining an active state can not be neglected when the number of participants increases drastically. This conclusion is ignored by most of the prior researchers. Second, accumulated consumed time decreases with the increase of sending interval, but maintaining an active state also consumes enormous time and energy in the CCS and CCS-fuzzy. In addition, the time consumed by DCS is less than CCS and CCS-fuzzy. The reasons are as follows. The vehicle-to-vehicle latency using DSRC communication is smaller than the latency from the vehicle to the cloud. Additionally, the multi-objective evaluator running on the local not only compresses the whole information but protected privacy by avoiding sending the whole information involving the vehicle to the neighbours. Finally, broadcasting evaluation is restricted in each small area, such as the range of 200 meters, and reduces the communication overhead.

\section{Conclusion}
\label{section7}
In this paper, we proposed a novel client selection scheme, namely distributed client selection, in which the FL server is not in charge of the client selection and does not gather information involving the participating vehicles. Furthermore, we proposed an evaluator with multi-objective, which run on each participating vehicle to obtain the evaluation of itself. In the evaluator, we considered four variables related to successfully uploading ratio and local dataset quality, specifically, sample quantity, throughput available, computational capability and loss function of the local dataset. Considering the non-existence of closed-form solutions over the four variables mentioned above, we developed fuzzy logic as the evaluator. Extensive simulations are conducted and verified the proposed scheme approached centralized client selection approximately. Meanwhile, the proposal cut down the communication overhead caused by maintaining the active state of all participating vehicles.
\section*{Compliance with Ethical Standards}

\textbf{Funding} This work was supported in part by Inner Mongolia autonomous region directly affiliated colleges and universities fundamental scientific research project (NCYWT23035).
\\
\\
\noindent \textbf{Disclosure of potential conflicts of interest} Narisu Cha declares that he has no conflict of interest. Long Chang declares that he has no conflict of interest.
\\
\\
\noindent \textbf{Research involving human participants and/or animals} This article does not contain any studies with human participants or animals performed by any of the authors. 
\\
\\
\noindent \textbf{Informed consent} Informed consent was obtained from all individual participants included in the study.
\\
\\
\noindent \textbf{Data availability} There is no any data availability for paper.
\\
\\
\noindent \textbf{Authors' contributions} All authors contributed equally to this work.


\bibliography{softcomputing}


\end{document}